\documentclass[letterpaper]{article} % DO NOT CHANGE THIS
\usepackage{aaai24}  % DO NOT CHANGE THIS
\usepackage{times}  % DO NOT CHANGE THIS
\usepackage{helvet}  % DO NOT CHANGE THIS
\usepackage{courier}  % DO NOT CHANGE THIS
\usepackage[hyphens]{url}  % DO NOT CHANGE THIS
\usepackage{graphicx} % DO NOT CHANGE THIS
\usepackage{natbib}  % DO NOT CHANGE THIS AND DO NOT ADD ANY OPTIONS TO IT
\usepackage{caption} % DO NOT CHANGE THIS AND DO NOT ADD ANY OPTIONS TO IT
\usepackage{algorithm}
\usepackage{algorithmic}

\usepackage{amsmath,amsfonts}
\usepackage{array}
\usepackage{textcomp}
\usepackage{url}
\usepackage{verbatim}
\usepackage{graphicx}
\usepackage{booktabs}
\usepackage{cite}
\usepackage{algorithm}
\usepackage{algorithmic}
\usepackage{diagbox}
\usepackage{multirow}
\usepackage{makecell}
\usepackage{engord}

\DeclareMathOperator{\sign}{sign}
\DeclareMathOperator{\ReLU}{ReLU}

\usepackage{newfloat}
\usepackage{listings}
\DeclareCaptionStyle{ruled}{labelfont=normalfont,labelsep=colon,strut=off} % DO NOT CHANGE THIS
\floatstyle{ruled}
\newfloat{listing}{tb}{lst}{}
\floatname{listing}{Listing}
\title{Analyzing Generalization in Policy Networks: A Case Study with the Double-Integrator System}
\author{
    Ruining Zhang\textsuperscript{\rm 1,2}\equalcontrib, 
    Haoran Han\textsuperscript{\rm 1}\equalcontrib, 
    Maolong Lv\textsuperscript{\rm 3}, 
    Qisong Yang\textsuperscript{\rm 4}, 
    Jian Cheng\textsuperscript{\rm 1}\thanks{Corresponding author.} \\ 
}
\affiliations{
	\textsuperscript{\rm 1}School of Information and Communication Engineering, University of Electronic Science and Technology of China\\
    \textsuperscript{\rm 2}Glasgow School, University of Electronic Science and Technology of China\\
    \textsuperscript{\rm 3}Air Traffic Control and Navigation College, Air Force Engineering University\\
    \textsuperscript{\rm 4}Xi'an Institute of High-Tech\\
    \{zhangruining0621, hanadam, maolonglv\}@163.com, q.yang@tutanota.com, chengjian@uestc.edu.cn
}
\usepackage{bibentry}

\begin{document}

\maketitle

\begin{abstract}
Extensive utilization of deep reinforcement learning (DRL) policy networks in diverse continuous control tasks has raised questions regarding performance degradation in expansive state spaces where the input state norm is larger than that in the training environment. This paper aims to uncover the underlying factors contributing to such performance deterioration when dealing with expanded state spaces, using a novel analysis technique known as state division. In contrast to prior approaches that employ state division merely as a post-hoc explanatory tool, our methodology delves into the intrinsic characteristics of DRL policy networks. Specifically, we demonstrate that the expansion of state space induces the activation function $\tanh$ to exhibit saturability, resulting in the transformation of the state division boundary from nonlinear to linear. Our analysis centers on the paradigm of the double-integrator system, revealing that this gradual shift towards linearity imparts a control behavior reminiscent of bang-bang control. However, the inherent linearity of the division boundary prevents the attainment of an ideal bang-bang control, thereby introducing unavoidable overshooting. Our experimental investigations, employing diverse RL algorithms, establish that this performance phenomenon stems from inherent attributes of the DRL policy network, remaining consistent across various optimization algorithms.
\end{abstract}

\section{Introduction}
Reinforcement learning (RL) stands as a pivotal technique for attaining artificial general intelligence. Its integration with deep learning, called DRL, empowers agents to effectively tackle intricate tasks in high-dimensional state space \cite{drl_intro, drl_intro2}. DRL has demonstrated remarkable performance in diverse gaming scenarios and exhibited promising potential in addressing real-world challenges \cite{drl_intro, drl_intro3}. Nevertheless, empirical observations indicate that DRL agents encounter difficulties in generalizing learned behaviors to environments with variants \cite{Ge_intro_2, Ge6, Ge_intro_1}. As the testing state space expands, a prevalent outcome is the performance degradation of DRL agents. For example, an agent proficiently navigating a $32\times32$ environment might struggle to reach the target in a $64\times64$ environment \cite{example3264}. Despite the ubiquity of this performance degradation and the existence of various proposed remedies, the root cause remains noninterpretable and has scarcely been discussed in prior literature.

Existing solutions to generalization problems can be classified into two categories \cite{packer2019assessing, hansen2021selfsupervised}. The first one revolves around fortifying the intrinsic robustness of policies \cite{robust_sz, robust_2, robust_3}. These learned policies can be directly applied in unfamiliar testing environments without updating. A prominent strategy to cultivate policies impervious to environmental changes is domain randomization, which exposes agents to diverse training environments endowed with random properties \cite{data_randomization_tobin, Domain_random_3, Domain_random_4, Domain_random_2}. While this always leads to suboptimal policies characterized by high variance, techniques such as regularization can alleviate these issues and enhance efficacy \cite{regularization_1, regularization_2}. Data augmentation has also demonstrated effectiveness in bolstering policy robustness \cite{augmented_data_1, augmented_data_2, augmented_data_3}. Specialized techniques such as adversarial data augmentation \cite{adversarial_aug_2, adversarial_aug_3, adversarial_aug_1} and $mixreg$ \cite{mixreg} excel in this realm. The second category of solutions entails dynamically adapting policies to accommodate various environments. When confronted with unseen environments, agents undergo retraining to update learned policies \cite{dynamic_adjust_2, dynamic_adjust_3, dynamic_adjust_1}. In cases where the reward function in testing environments is readily available, supervised learning proves advantageous for swiftly fine-tuning existing policies \cite{finetune_1, finetune_3, finetune_2}. However, formulating an appropriate reward function often remains infeasible \cite{hansen2021selfsupervised}. As an alternative, self-supervised learning is introduced to enhance the generalization of DRL policies through optimizing representation learning \cite{self_representation_1, Ge6, self_representation_sz}. For example, \citet{hansen2021selfsupervised} devise a joint-objective system that integrates a self-supervised objective to enhance intermediate representation optimization through self-supervised learning. 

Apart from supervised and self-supervised learning, approaches such as meta-learning \cite{meta1Ge2, meta2}, transfer learning \cite{transfer2, transfer1} and curriculum learning \cite{curriculum_1, curriculum_2, curriculum_3} have also demonstrated their capacity to adapt pre-trained policies to novel environments through retraining \cite{packer2019assessing}. Integration of techniques such safe RL is recommended within the training process to allow the agent to avoid hazardous states, thereby ensuring consistent normal performance and enhancing training efficiency \cite{safe1, safe2}. When confronted with expanded state space, the latter category of solutions that retrain learned policies is often favored. However, crafting and operating such an extensive state space proves inefficient and resource-intensive.

The application of state division serves as an intuitive and powerful method for interpreting DRL policies, which currently functions as an auxiliary tool for mimic learning and explicating policy performance. In the context of mimic learning, \cite{Soares2020_Inter1_2, Dhebar2022_Inter1_1, Liu2023_Inter1_3} visualize state-action patterns to refine approximations. For instance, \cite{Liu2023_Inter1_3} define critical experience points around decision boundaries, establishing their significance in interpretable policy distillation. In contrast, \cite{pascanu2014_count2, serra2018_count1, hanin2019_node, divi_main} delve deeper into the origins and potential impact of state division, focusing on the piecewise linear mapping from states to actions induced by activation in each node. \cite{pascanu2014_count2, serra2018_count1, hanin2019_node} build theories concerning on linear region counting and numbers, based on which \cite{divi_main} investigate the relationship between region density along the policy trajectory and policy performance. Notably, both applications of state division solely provide post-hoc explanations for currently learned policies \cite{DT_1,post_hoc_2}, as division lines exhibit irregular variations across different cases. They lack generalizability to an expanded state space and diverse input conditions.

To enhance the policy network generalization in expanded state space, it is imperative to unravel the underlying mechanisms of networks. While state division serves as an intuitive tool, prior works predominantly employ it for emulating noninterpretable policies or treat each piecewise region uniformly, which offers a constrained and superficial perspective. In contrast, our approach harnesses state division to delve into the intrinsic properties of the policy network. This enables us to assess the influence of these properties on network generalization. In contrast, our approach focuses on identifying a siginificant division line in the state space and harness state division to delve into the intrinsic properties of the policy network. This enables us to assess the influence of these properties on network generalization.

This paper introduces state space division theory as a means to interpret DRL policies. The essence of this theory lies in the continuous division line in the state space, whose configuration evolves with varying state norms. While the resulting policy remains nonlinear, distince patterns emerge in the division in the state space division. As shown in Fig. 1, in the proximal state space where the state norm is small, this division line exhibits a nonlinear nature owing to the unsaturated activation function $\tanh$. As the state norm approaches infinity, the policy network approaches saturation, resulting in a predominantly linear division line in the infinite state space. Moreover, accommodating the context of a continuous action space, this division line transforms into a division strip, within which the network outputs are continuous. In the finite distal state space, closely positioned division strips overlap, encompassing a radial boundary. To empirically illustrate these principles, we leverage the double-integrator system as an example, validating that network performance degradation, such as overshoot, stems from the nonlinear decline in policy networks. The source code is available at https://github.com/Han-Adam/GeneralAnalyze.

\begin{figure}[!t]
	\centering
	\includegraphics[]{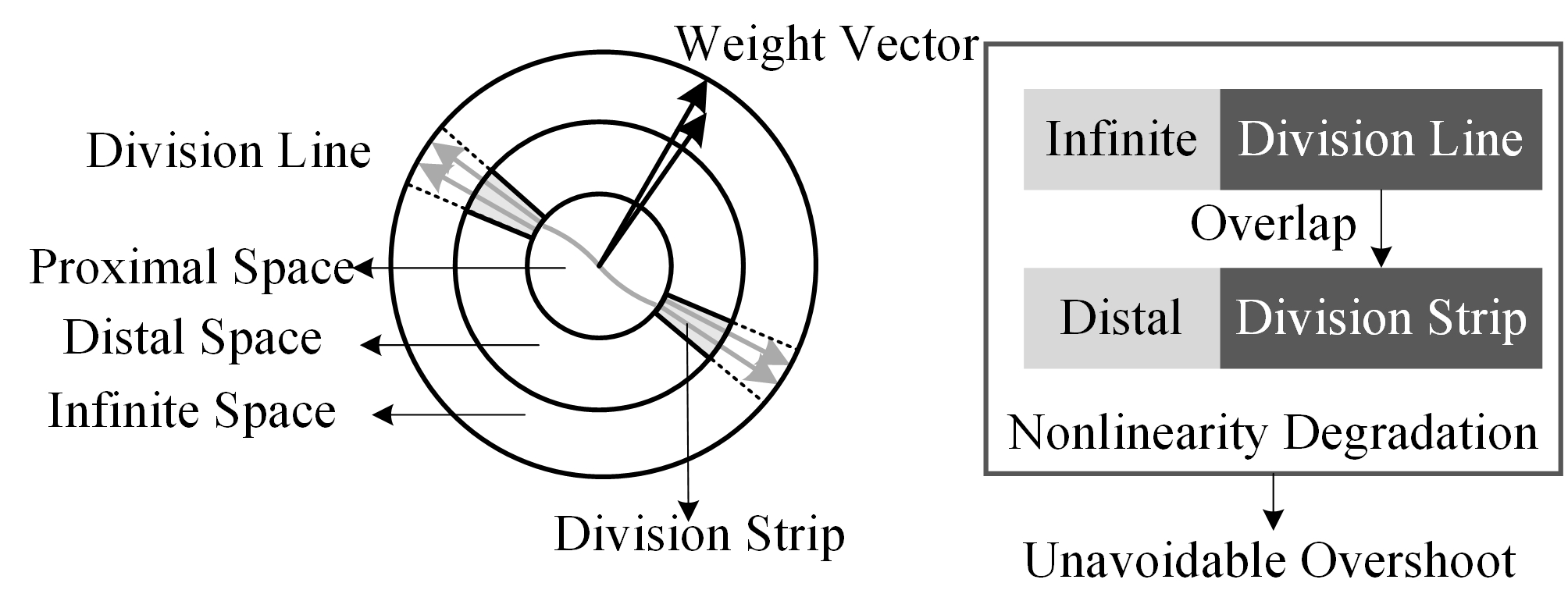}
	\caption{Overview of the state space division theory and article structure. }
\end{figure}

\section{Background}
\subsection{Deep Reinforcement Learning}
The mathematical formulation of the RL problem is the Markov decision process (MDP) described as $(\mathcal{S}, \mathcal{A}, \mathcal{P}, R, \gamma)$. The agent receives the state $s_t\in\mathcal{S}$, and takes an action $a_t\in\mathcal{A}$. Then, the environment generates the next state $s_{t+1}\sim \mathcal{P}(\cdot|s_t,a_t)$ and the reward $r_t=R(s_t)$. For any given trajectory, the discounted reward is defined as $G_t = \sum_{i=0}^\infty\gamma^ir_{t+i}$. The objective of an RL algorithm is to train its policy $\mu$ to maximize the discounted reward the agent can obtain.

\subsection{Network Based Control}
DRL adopts neural networks, whose parameters are denoted as $\theta$, for policy approximation. In this paper, the MLP, a concatenation of $m$ fully connected layers of 
\begin{equation}
	\label{fullyConnect}
	z^i = \sigma(W^iz^{i-1} + b^i) \\
\end{equation}
is adopted. Here, $i=1,...,m$ means the layer index, $W^i\in\mathbb{R}^{n_{i}\times n_{i-1}}$ and $b^i\in\mathbb{R}^{n_i}$ are the weight matrix and bias, $n_i$ is the width of the feature $z^i$, and $\sigma$ is the activation function. In particular, $z^0=s$ is the network input. The weight matrix could be expanded as $W^i = [w^{iT}_1, w^{iT}_2, ..., w^{iT}_{n_i}]^T$, where $w^i_j\in\mathbb{R}^{n_{i-1}}$ is the weight vector.

Previous works introduce a simplified policy network, where the bias $b^i$ in Eq. \eqref{fullyConnect} is eliminated and the activation function should satisfy the condition $\sigma(0)=0$. This approach has demonstrated enhanced efficiency in attaining stable control performance, as proven through both the Lyapunov method \cite{simplified_1, simplified_2} and an analysis of system linearization \cite {simplified_han}. Therefore, we leverage the simplified policy network in our investigation, whose fully connected layer should be rewritten as
\begin{equation}
	z^i = \tanh(W^iz^{i-1}).
\end{equation}

\subsection{Double-Integrator}
The double-integrator is described as
\begin{equation}
	\label{doubleIntegrator}
	\dot{p} = v,
	\dot{v} = a ,
\end{equation}
where $p\in\mathbb{R}$, $v\in\mathbb{R}$, and $a\in[-\bar{a}, \bar{a}]$ respectively represent the position, velocity and acceleration of the agent. $\bar{a}$ is the acceleration bound. Without losing generality, this paper sets $\bar{a}=5$ for the later statement.

% In addition, the linearization coefficient can be calculated as $k_\theta=W^lW^{l-1}...W^1$ if we apply the odd policy network, which provides an analytical way to check whether the stable control performance could be achieved.

\section{State Space Division}
\subsection{Division Line}
If the state is a nonzero vector, it can be expressed as $s = dl$, where $l = ||s||$ is the length of the state and $d=s/l$ is the direction of the state. To demonstrate the linear division of the policy network, we define the function
\begin{equation}
	\phi(d) = \lim_{l\rightarrow\infty} z^1
\end{equation}
as the feature vector of the first layer fed with an infinite state. Here, the domain of $\phi$ is the unit circle of 
$\mathcal{C} = \{d\in\mathbb{R}^2: ||d||=1\}$.
This function can be expanded as
\begin{equation}
\begin{aligned}
	\phi(d) &= \lim_{l\rightarrow\infty} \tanh(W^1dl)\\
	& =\lim_{l\rightarrow\infty} [\tanh(w_1^{1T}d l ), ..., \tanh(w_{n_1}^{1T}d l )]^T.
\end{aligned}
\end{equation} 
Moreover, each element of $\phi(d)$ is calculated as
\begin{equation}
	\phi_i (d)= \left\{
	\begin{matrix}
		-1, & \text{if } w_{i}^{1T}d < 0.\\
		0, & \text{if } w_{i}^{1T}d = 0.\\
		1, & \text{if } w_{i}^{1T}d > 0.
	\end{matrix}
	\right.
\end{equation}
This implies that, when the norm of the state goes to infinity, the first layer of the policy network encodes the state into a permutation consisting of $-1$, $0$, and $1$. Thus, $\phi$ is a piecewise-constant function, and the discontinuity occurs if there exist weight vectors $w^1_i$ such that $w^1_i\perp d$. 

Then, we can define the division direction set associated with the weight vector $w^1_i$ as 
$\mathcal{D}_i = \{d\in\mathcal{C}: d\perp w_i^1 \}$.
These division directions divide the unit circle $\mathcal{C}$ into multiple regions, where each region is associated with a constant feature vector. Therefore, for any given representative direction $d$ that is not perpendicular to any weight vectors, the division region expanded from it can be defined as
$\mathcal{R}(d)  = \{d' \in \mathcal{C}: \phi(d') = \phi(d) \}$.

Additionally, for the sake of simplicity, this paper assumes that no weight vector is parallel to another.
With this assumption, the features in any two adjacent regions have only one different element. Concretely, for any two adjacent regions $\mathcal{R}(d_1)$, $\mathcal{R}(d_2)$, where $d_1$ and $d_2$ are the representative directions, if the direction $d\in\mathcal{D}_i$ is their boundary, one can obtain that $\phi_i(d_1) = -\phi_i(d_2)$ and $\phi_j(d_1) = \phi_j(d_2)$ for all $j\neq i$, such as that in Fig. 2.

\begin{figure}[!t]
	\centering
	\includegraphics[]{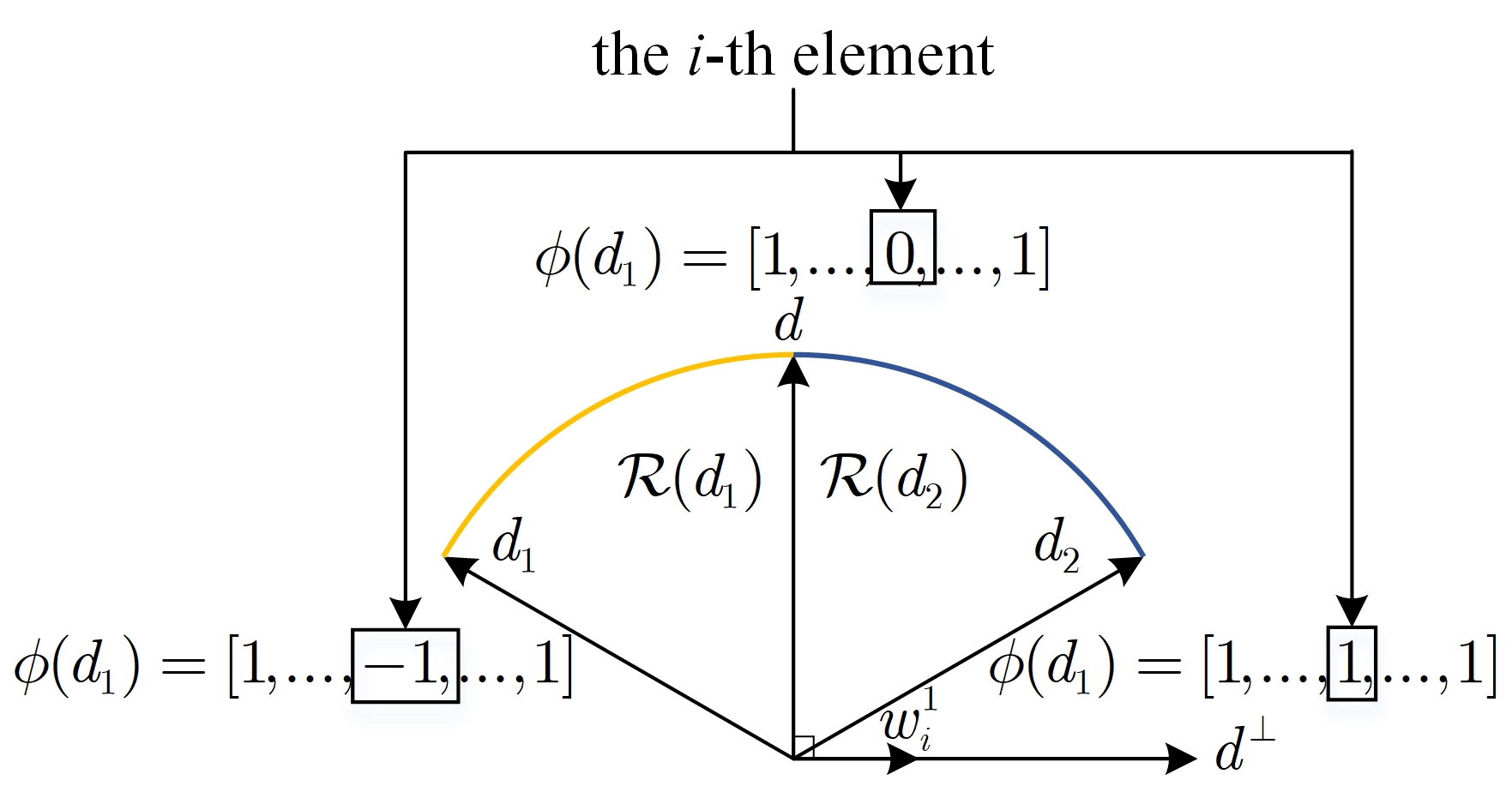}
	\caption{Illustration of $\phi$ among adjacent regions.}%  The main direction $d$ is perpendicular to the weight vector $w^1_i$. Only the $i$-th element of the function $\phi$ changes as the input direction changes.}
\end{figure}

If the directions of two states $s$ and $s'$ fall into the same region and their norms go to infinity, the first layer of the policy network will encode them with the same feature. 
Therefore, the ultimate output of the policy network, 
\begin{equation}
	\bar{\phi}(d) = \lim_{l\rightarrow\infty} \bar{a}\mu_\theta(s) = \bar{a}\tanh(...\tanh(W^2\phi(d))...),
\end{equation}
is also divided by these division directions.

For a real policy network, there are nonlinear division boundaries in the proximal state space. That is because $\tanh$ is far from reaching saturation when the state norm is small. As the state norm grows, $\tanh$ activation gradually tends to saturate, and the feature $z^1$ converges to $\phi(d)$. Therefore, the division direction defined on the unit circle can be generalized to the division line defined on the whole space as
\begin{equation}
	\bar{\mathcal{D}}_i = \{s\in \mathcal{S}:s\perp w_i^1 \}.
\end{equation}

\subsection{Division Strip}
Since the policy network is a continuous function, it is inadequate to characterize its behavior using discontinuous division lines. More realistically, the policy network generates division strips around these lines, indicating that the output changes continuously as the state transforms between adjacent regions. Hence, we define the strip function as
\begin{equation}
	\psi(d, d^\perp, x) = \lim_{l\rightarrow\infty} z^1 = \lim_{l\rightarrow\infty} \tanh(W^1(dl+d^\perp x)),
\end{equation}
where $d^\perp$ is perpendicular to $d$. The function $\psi$ assumes the state extends infinitely in the main direction $d$ but has a finite offset of $x$ in the direction of $d^\perp$.
For a weight vector such that $w_{i}^{1T}d \neq 0$, the offset has no effect on the output since $\lim_{l\rightarrow\infty}w^{1T}_i(dl+d^\perp x) =\sign(w_{i}^{1T}d) \infty + w^{1T}_id^\perp x = \sign(w_{i}^{1T}d)\infty$.
Without loss generality, this paper assumes $w^{1T}_id^\perp > 0$ and each element of $\psi$ can be calculated as
\begin{equation}
	\psi_i (d, d^\perp, x)= \left\{
	\begin{matrix}
		-1, & \text{if } w_{i}^{1T}d < 0.\\
		\tanh(||w^1_i|| x), & \text{if } w_{i}^{1T}d = 0.\\
		1, & \text{if } w_{i}^{1T}d > 0.
	\end{matrix}
	\right.
\end{equation}

When the main direction $d$ is not perpendicular to any weight vector, the function $\psi$ is equal to $\phi$. That means the offset has no effect on the feature if the main direction is inside a division region. When the main direction $d$ is a division direction, without losing generality, we assume that $w^{1T}_i d_1 > 0$, $w^{1T}_i d_2 < 0$, $\sign(w^{1T}_jd_1) = \sign(w^{1T}_jd_2)=\sign(w^{1T}_jd)$ for all $j\neq i $, and $d_1$ and $d_2$ are the representative vectors of the adjacent regions, $\mathcal{R}(d_1)$ and $\mathcal{R}(d_2)$, where $d$ is their boundary, such as that in Fig. 2.
Hence, $\psi_j(d,d^\perp,x) = \phi_j(d_1) = \phi_j(d_2)$ for all $j \neq i$, and 
\begin{equation}
	\begin{aligned}
		\lim_{x\rightarrow\infty} \psi_i (d, d^\perp, x) & = 1 = \phi_i(d_1), \\
		\lim_{x\rightarrow-\infty} \psi_i (d, d^\perp, x) & = -1= \phi_i(d_2).
	\end{aligned}
\end{equation}
That implies that $\psi(d, d^\perp, x)$ converges to $\phi(d_1)$ or $\phi(d_2)$ as the offset $x$ grows.
Considering that $\psi(d,d^\perp,0)=\phi(d)$, the function shows how the feature signal $z^1$ gradually changes among the adjacent division regions.

Consequently, the ultimate output of the policy network,
\begin{equation}
	\begin{aligned}
		\bar{\psi}(d, d^\perp, x) & = \lim_{l\rightarrow \infty} \bar{a}\mu_\theta (s) \\
		&=  \bar{a}\tanh(...\tanh(W^2\psi(d, d^\perp, x))...)
	\end{aligned}
\end{equation}
also gradually changes. Therefore, the policy network generates a strip around the division line $\bar{\mathcal{D}}_i$, and the network output gradually changes within this strip.

In infinite space, it can be asserted that the alteration in the direction of the state leads to the modification of solely one element in the feature of the first layer $z^1$, at any given moment.
However, these strips and adjacent division regions may overlap in the finite distal state space. Specifically, if the directions of a cluster of weight vectors are in proximity, their corresponding division lines will also be closely positioned. Hence, the regions demarcated by these lines become considerably narrower. Therefore, these strips and regions collectively form an overlapped division strip wherein multiple elements of $z_1$ alter as the state direction changes.  

\subsection{Weight Vector Significance}
The division line $\bar{\mathcal{D}}_i$ is perpendicular to the weight vector $w^1_i$. However, not all division lines contribute equally to the network performance. If we assume the direction $d$ is perpendicular to $w^1_i$, and denote $\delta_i = \tanh(||w^1_i|| x)$ as the offset on the feature vector, then the strip function of Eq. (9) can be rewritten as $\psi(d,\delta_i)$, where
\begin{equation}
	\psi_j (d, \delta_i)= \left\{
	\begin{matrix}
		-1, & \text{if } i\neq j \wedge w_{i}^{1T}d < 0.\\
		\delta_i, & \text{if } i=j.\\
		1, & \text{if } i\neq j \wedge w_{i}^{1T}d > 0.
	\end{matrix}
	\right.
\end{equation}
Then, the ultimate network output can be rewritten as  
\begin{equation}
	\bar{\psi} (d, \delta_i)= \tanh(...\tanh(W^2\psi(d,\delta_i))...),
\end{equation}
and the significance of the division line $\bar{D}_i$ is defined as
\begin{equation}
	\rho_i = |\bar{\psi} (d, 1) - \bar{\psi}(d, -1)|.
\end{equation}

Noticing that $\bar{\psi} (d, 1)$ and $\bar{\psi} (d, -1)$ are the network outputs of two adjacent division regions, the significance of the weight vector indicates the difference of the network outputs when the state transforms between adjacent regions. Thus, significant disparities across the line imply noticeable changes in network output.

\begin{figure}[!t]
	\centering
	\includegraphics[]{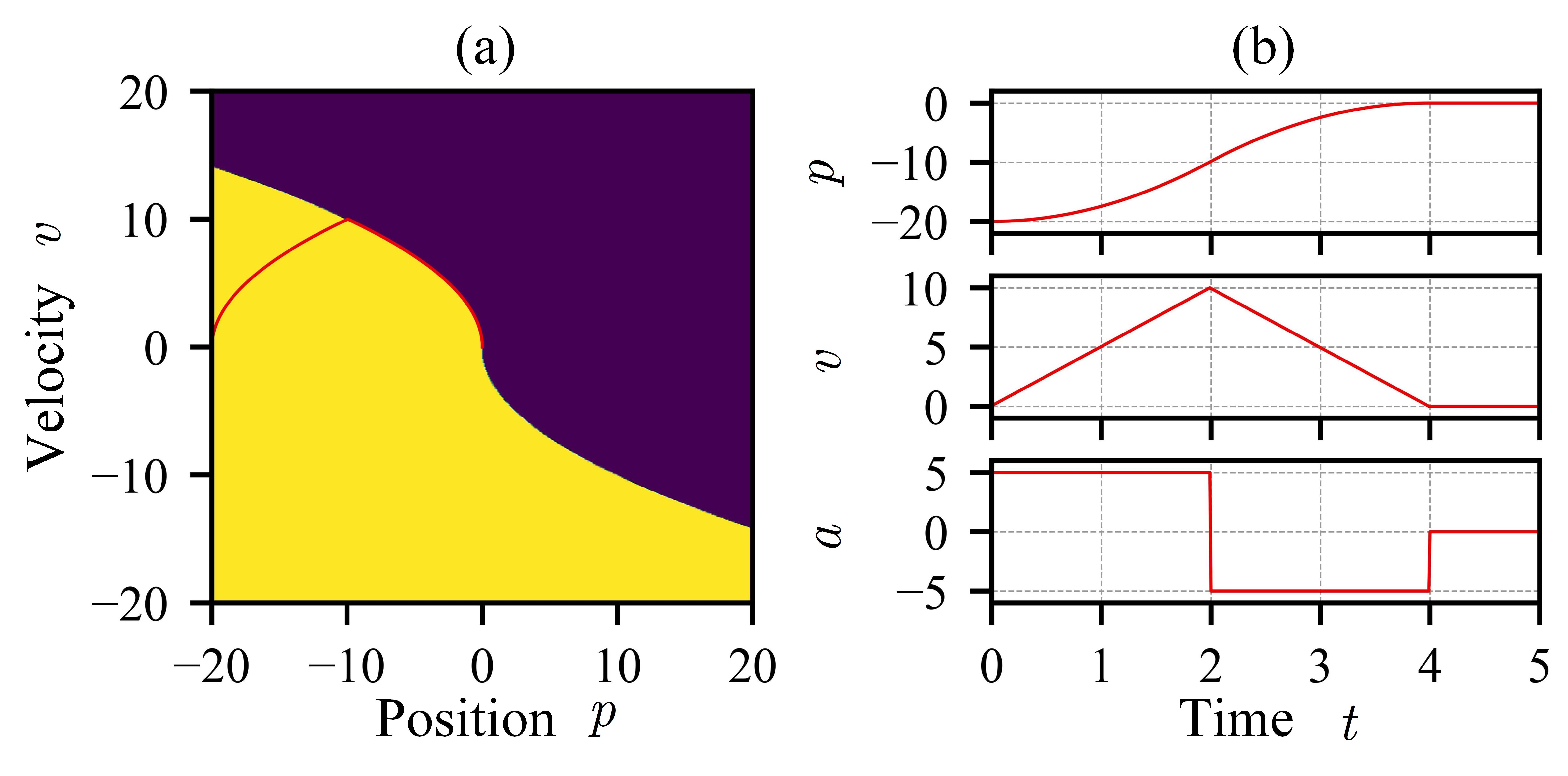}
	\caption{(a) State trajectory (red) and state-action pattern of the ideal bang-bang control. (b) Response of $p$, $v$, and $a$ via time under the control law of Eq. (17).}
\end{figure}

\subsection{Unavoidable Overshoot}
According to the preceding discussion, it is observed that as the input state tends towards infinity, the policy network generates a division line that separates the state space into two regions, wherein the acceleration output approximates either $-\bar{a}$ or $\bar{a}$. This control scheme bears resemblance to a bang-bang control. Nevertheless, an ideal analytic bang-bang control exhibits a division line denoted as

\begin{equation}
	\label{idealNonlinear}
	2\bar{a}p = -\sign(v)v^2.
\end{equation}
Thus, its output acceleration can be represented as
\begin{equation}
	a= \left\{
	\begin{matrix}
		-\bar{a}, & \text{if }2\bar{a}p>-\sign(v)v^2,\\
		0, & \text{if }2\bar{a}p=-\sign(v)v^2,\\
		\bar{a}, & \text{if }2\bar{a}p<-\sign(v)v^2,\\
	\end{matrix}
	\right.
\end{equation}and its state-action pattern is shown in Fig. 3 (a). This particular control algorithm is time-optimal \cite{bang_bang_control} and can attain non-overshoot control performance. However, the policy network is limited in generating a linear division line, thereby unable to approximate the optimal division line specified in Eq. (16). Consequently, time-optimal control cannot be achieved using the policy network. When the initial position error is relatively large, the agent will commence decelerating later%, as exemplified by the situation illustrated in Fig. 5 (a)
, leading to unavoidable overshoots.

Subsequently, we assert that if the policy network confronts considerable initial error, the occurrence  of overshoot is unavoidable. This outcome is not primarily attributed to differences in optimization objects (DRL optimizes the discounting reward rather than the time consumption), but is an inherent consequence of the network saturability.

\section{Experiments}

\begin{figure}[!t]
\centering
\includegraphics[]{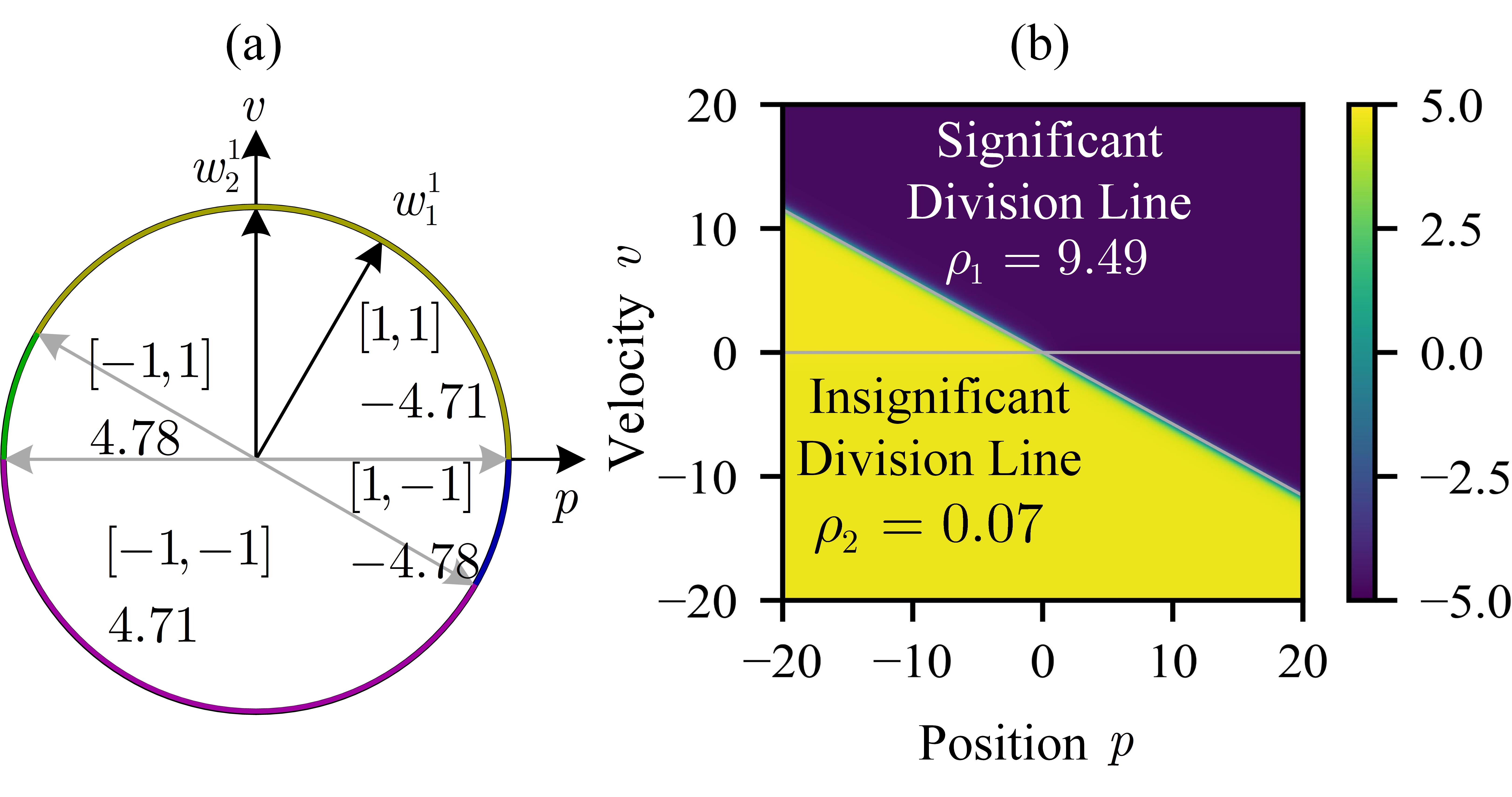}
\caption{(a) Unit circle divided by the division directions perpendicular to the weight vector. (b) State-action pattern divided by two division lines.}
\end{figure}

\begin{figure}[!t]
\centering
\includegraphics[]{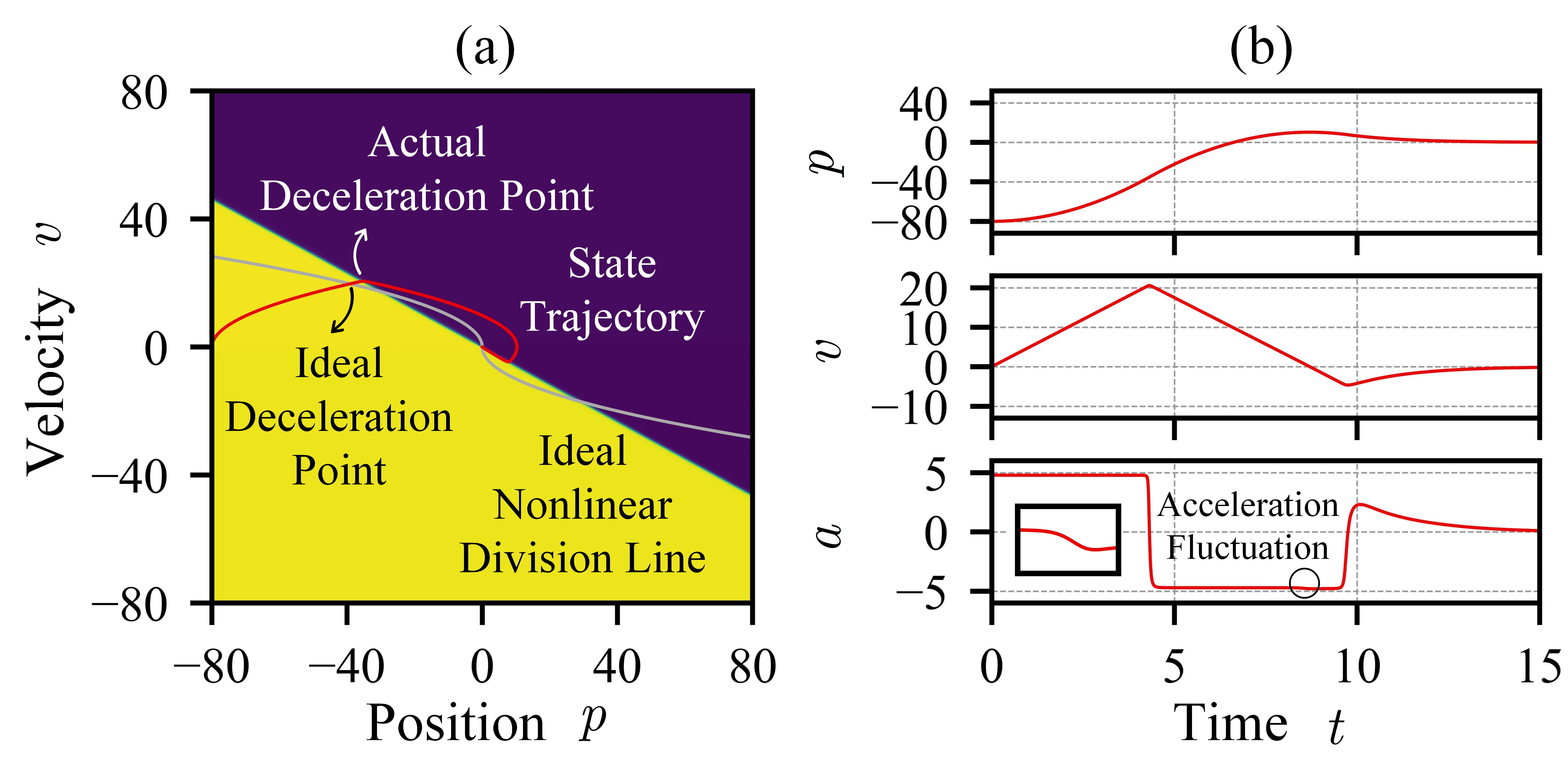}
\caption{(a) The state trajectory (red) and ideal nonlinear division line (gray). (b) The response of $p$, $v$, and $a$ via time.}
\end{figure}

\subsection{Artificially Constructed Examples}
To demonstrate how the weight vector and its significance characterize the network performance, we construct a network with parameters of

\begin{equation}
W^1= \left[
\begin{matrix}
\frac{1}{2} & \frac{\sqrt{3}}{2} \\
0 & 1\\
\end{matrix}
\right],
W^2 = \left[
\begin{matrix}
2 & -1\\
2 & \frac{1}{2}\\
\end{matrix}
\right],
W^3 = \left[
\begin{matrix}
-1\\
-1\\
\end{matrix}
\right].
\end{equation}
As shown in Fig. 4 (a), the unit circle is divided by $\mathcal{D}_1=\{[\sqrt{3}/2, -1/2]^T,[-\sqrt{3}/2, 1/2]^T\}$ and $\mathcal{D}_2=\{[1,0]^T,[-1,0]^T\}$
into four regions, each of which is assigned a different $\phi$ and $\psi$. However, as shown in Fig. 4 (b), only the division line of $\bar{\mathcal{D}}_1=\{s: s=[\sqrt{3}/2, -1/2]^Tl, l\in\mathbb{R}\}$ is significant. The influence of $\bar{\mathcal{D}}_2=\{s:v=0\}$ on the action is unobservable.

Then, we apply the exemplary network to accomplish the control task. The state trajectory and the responses of $p$, $v$, and $a$ are shown in Fig. 5. The agent first accelerates and then decelerates after its state crosses $\bar{D}_1$. The acceleration has a slight fluctuation of $0.07$ at $t=8.71$, where the velocity becomes zero and the state crosses the $\bar{D}_2$. This fluctuation is negligible and has little impact on performance.

Therefore, the direction of division lines is determined by the weight vectors of the first layer while the magnitude of the significance is dependent on subsequent layer parameters. Only those significant division lines determine the performance of the policy network.

\begin{figure}[!t]
	\centering
	\includegraphics[]{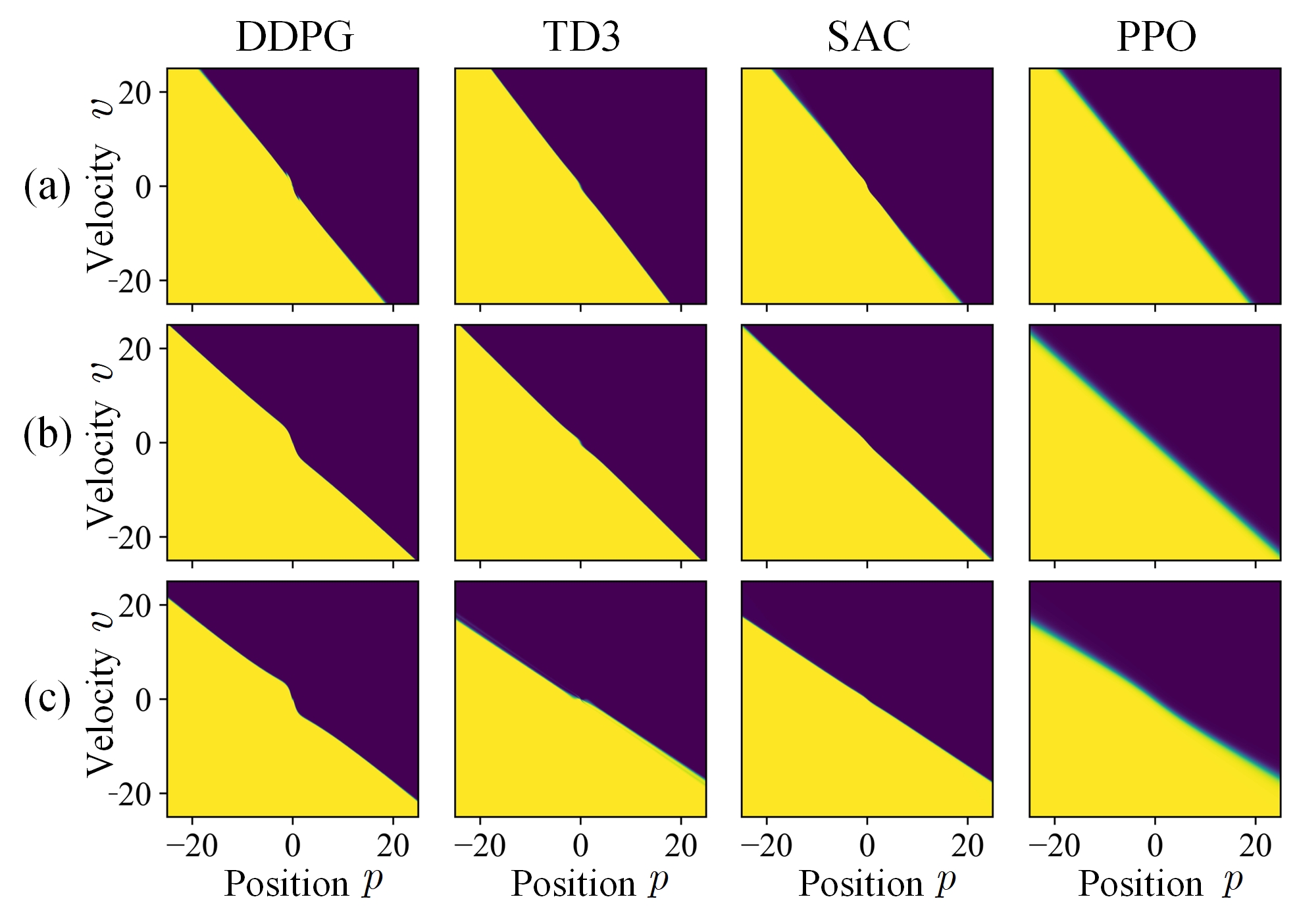}
	\caption{State-action patterns of the best-performing agents using different algorithms.}
\end{figure}

\begin{figure}[!t]
	\centering
	\includegraphics[]{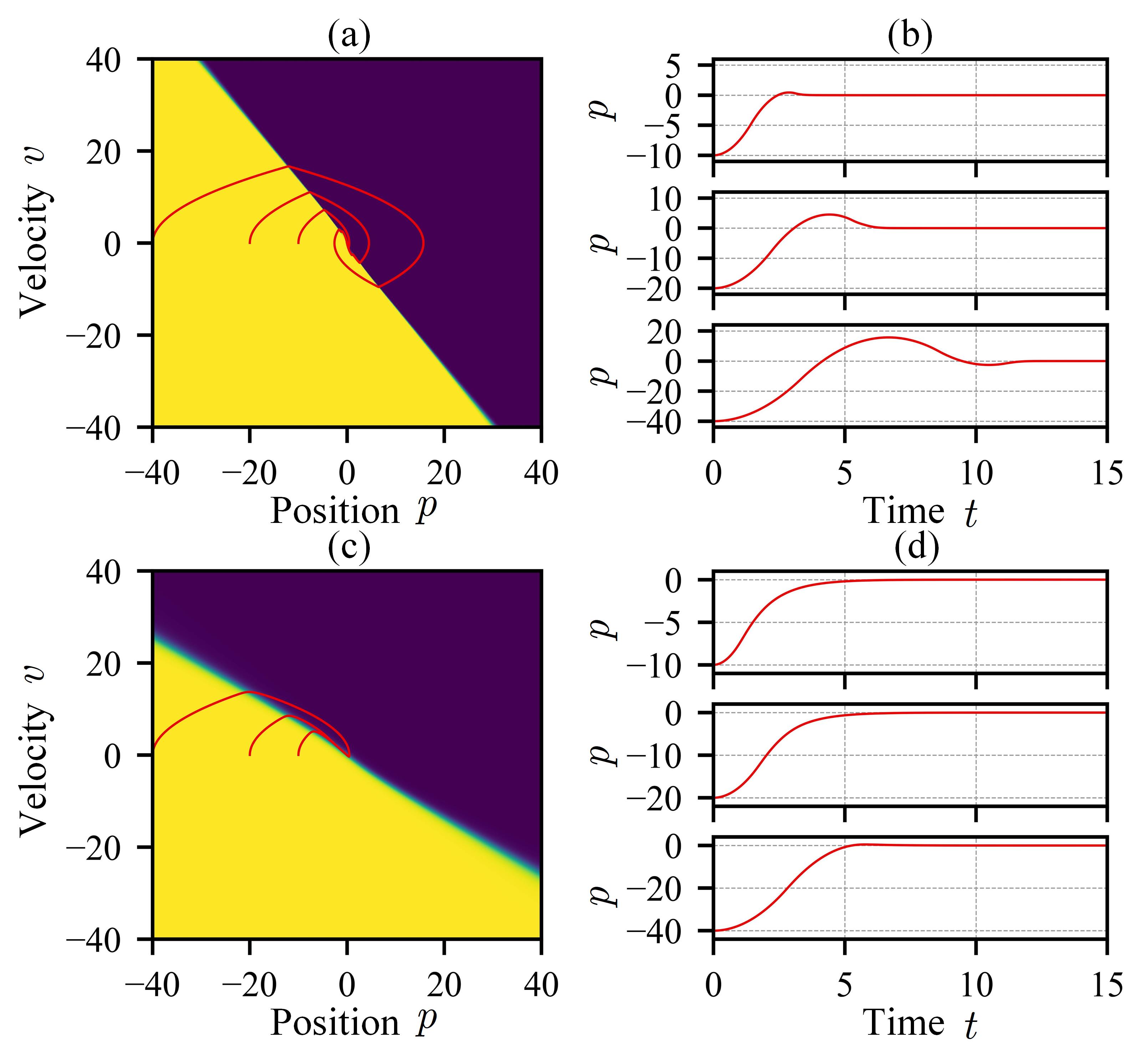}
	\caption{Performance of the network with different slopes of the division line facing different test cases.}
\end{figure}

\begin{figure*}[!t]
	\centering
	\includegraphics[]{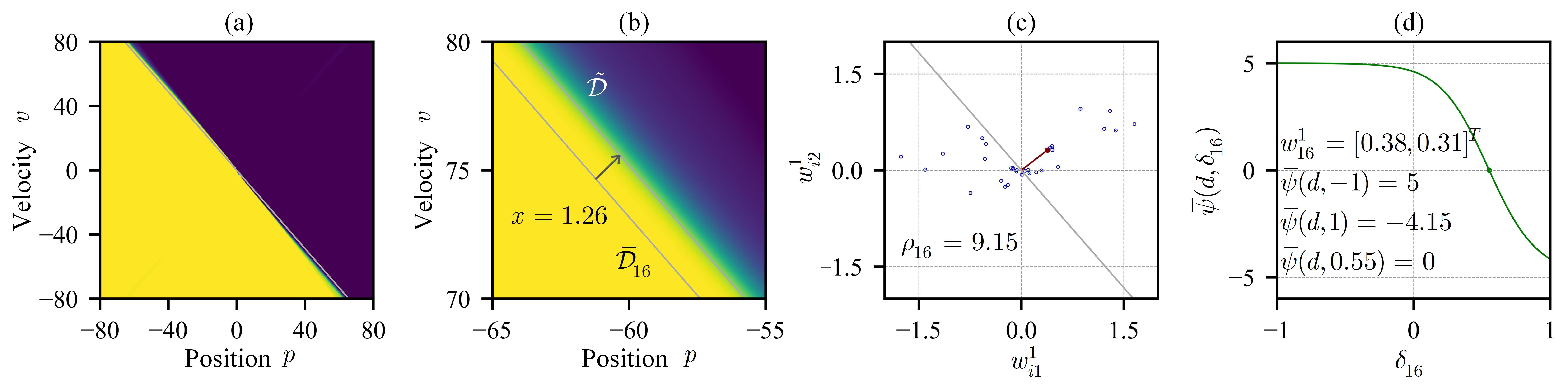}
	\caption{Analysis of network in DDPG (a) in Fig. 6, where $\tilde{\mathcal{D}} = \{s\in\mathcal{S}: \mu_\theta(s)=0\}$ and $\bar{\mathcal{D}}_{16} = \{s\in \mathcal{S}:s\perp w_{16}^1 \}$.}
\end{figure*}

\begin{figure*}[!t]
	\centering
	\includegraphics[]{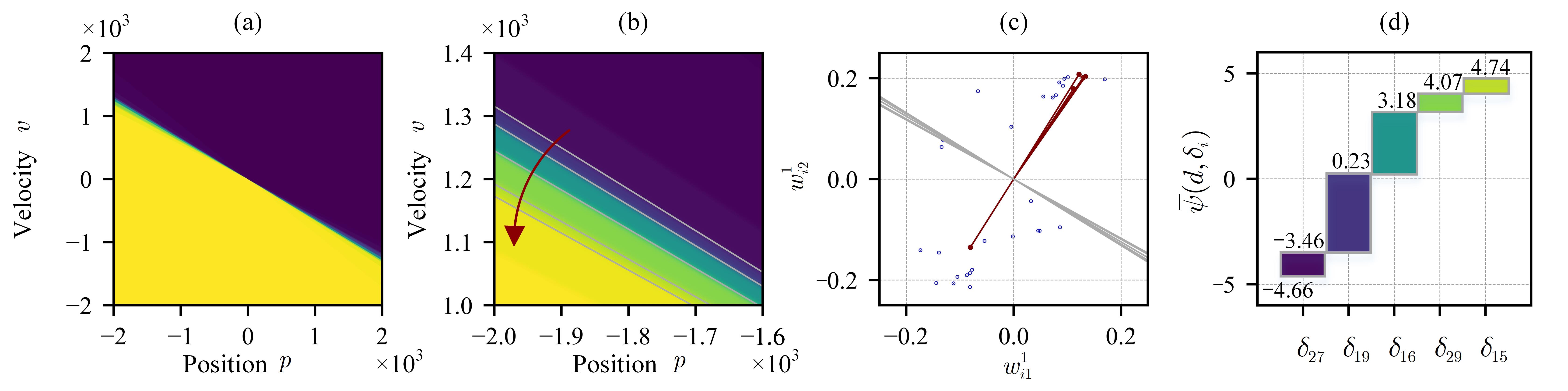}
	\caption{Analysis of network in PPO (c) in Fig. 6. }
\end{figure*}

\subsection{Examples Trained with Realistic Conditions}

In the subsequent section, we analyze experimental results derived from networks trained under realstic conditions to validate the theory regarding linear state space division and its implications for the generalization of network controller. The neural network employed in experiments is a simplified policy network comprised of three fully connected layers, each containing 32 nodes. To ensure the broad applicability of our experiments, four prominent DRL algorithms, DDPG \cite{DDPG}, TD3 \cite{TD3}, SAC \cite{SAC} and PPO \cite{PPO} are selected as examples.

\subsubsection{Universality of Linear Division Line}
The state-action patterns of the best-performing agent are shown in Fig. 6. In all cases, the initial velocity is set to zero and the initial position is set to $-10, -20$ and $-40$. A division line is evident in all testing cases, and it exhibits a nonlinear nature in the proximal state space, where the activation function $\tanh$ is unsaturated. As the state norm grows, $\tanh$ saturates, and these division lines become essentially linear when observed from the whole state space. Consequently, this linear division of the state space is a universal phenomenon that is determined by the intrinsic structure of the neural network and is irrelevant to the optimization algorithm.

\subsubsection{Impact of Linear Division}
Another notable phenomenon is the flattening of division lines as the initial position increases. It is analyzed by examining the state-action pattern of a general agent that does not exhibit optimal performance. As depicted in Fig. 5 (a), the state trajectory interacts with both the ideal division line and the actual division line, with their corresponding intersections defined as the ideal deceleration point and the actual deceleration point. The alignment of these two points indicates that the timing of deceleration for an agent is ideal, enabling it to achieve optimal performance. Hence, the division lines for the best-performing cases illustrated in Fig. 6 should intersect both the ideal deceleration point and the origin. Additionally, based on the characteristics of ideal bang-bang control, when the initial velocity is set to zero, the abscissa of the ideal deceleration point should be half the initial position error. Therefore, the ideal deceleration point moves away from the origin along the ideal division line with increasing initial position error, causing a flattening of the actual division line.

The following examples are evaluated to substantiate the aforementioned analysis. Figs. 7 (a) and (c) present DDPG (a) and PPO (c) in Fig. 6 with visualized state trajectories starting at $-10, -20$ and $-40$. Figs. 7 (b) and (d) display the corresponding position responses. In the context of a steeper division line, the agent starting at $-5$ swiftly reaches the origin, approximating optimal performance. However, agents beginning at $-20$ and $-40$ both exhibit late deceleration, resulting in noticeable overshoot. Conversely, for a flatter division line, the agent starting at $-40$ approximates its optimal performance, while those starting at $-20$ and $-10$ gradually approach the origin, demonstrating conservative behavior.

\subsubsection{Examples of Division Strips} This analysis examines cases related to division strips. Fig. 8 involves a policy network where $\bar{\mathcal{D}}_{16}$ represents the perpendiculars of the weight vector $w^1_{16}=[0.38, 0.31]^T$, bearing the highest significance of $\rho = 9.15$. As shown in Fig. 8 (b), the practical division line $\tilde{\mathcal{D}} = \{s\in\mathcal{S}: \mu_\theta(s)=0\}$ deviates from $\bar{\mathcal{D}}_{16}$ by $x = 1.26$ in the perpendicular direction of $\bar{\mathcal{D}}_{16}$. This deviation causes an offset $\delta_{16}$ on the feature weight vector, computed as $0.55$ using the strip function $\delta_{i}=\tanh(||w^1_{i}|| x)$. Notably, $\delta_{16}=0.55$ precisely corresponds to $\bar{\psi} (d, \delta_{16}) = 0$ in Fig. 8 (d). This implies that altering the direction of a state vector can induce an offset $\delta_{i}$ on the feature weight vector, ultimately influencing the network output.

An intriguing anomaly is observed in Fig. 9 (a), where the linear division line assumes a radial and blurry configuration. This phenomenon arises due to the existence of five closely positioned weight vectors, all of which bear comparable significance. These weight vectors generate five perpendicular division lines, as shown in Fig. 9 (c). Upon closer examination in Fig. 9 (b), the state-action patterns in the distal state space disclose a divergence of the division line cluster, manifesting in a substantial inter-line distance. As the state norm decreases, these strips mutually overlap with the progressive convergence of division lines, leading to the formation of a radial division boundary. In Fig. 9 (c), the network output within these strips is visualized. As state changes in the direction shown in Fig. 9 (b), the output exhibits a noticeable variation across each strip, which corresponds to their significance.

\begin{figure}[!t]
	\centering
	\includegraphics[]{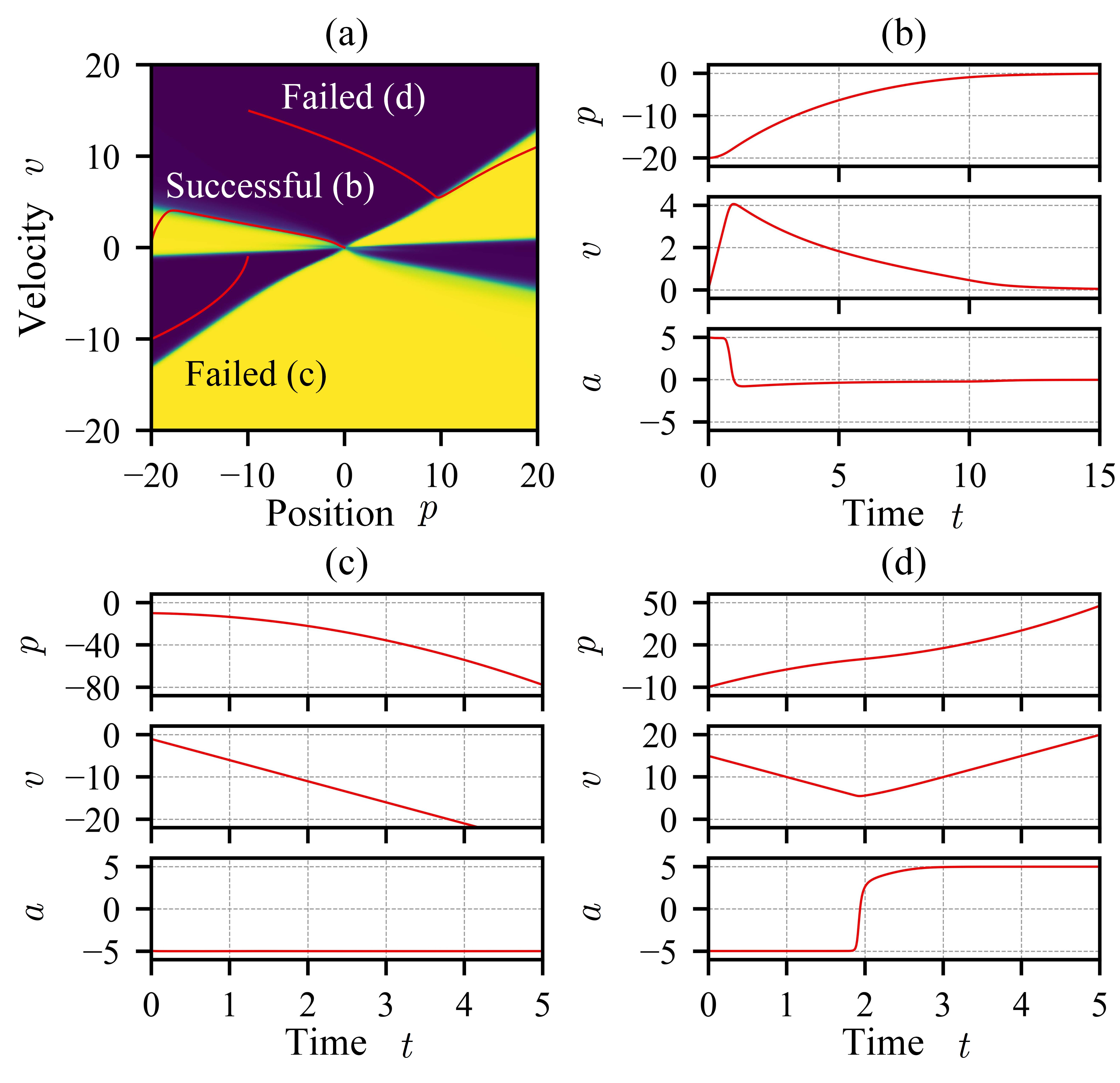}
	\caption{Illustration of a failed policy network. Dead zones exist in the state-action pattern (a). (b) showcases the response of a successful case, while (c) and (d) present the failed ones.}
\end{figure}

\subsubsection{Failed Policy Network}
Furthermore, an unsuccessful policy network is examined to ascertain the reason for its failure. As depicted in Fig. 10 (a), these state trajectories predominantly align with division lines, with some diverging and failing to converge at the origin. This can be attributed to the fact that the state of these agents falls within a dead zone created by radial division lines. Fig. 10 (a) shows that trajectories (d) and (c) both enter dead zones, where the position, velocity and generated acceleration share the same sign. Additionally, this dead zone uninterruptedly expands according to the definition of the division region $\mathcal{R}(d)$. Hence, once an agent enters a dead zone, its deviation from the target accelerates, leading to a permanent failure to reach it. Therefore, it is confidently asserted that the presence of a dead zone in the state space is the primary catalyst for the failure of a policy network. 

\subsubsection{General Policy Network}
To simplify our exploration, we have thus far focused on the performance of a simplified policy network employing $\tanh$ as the activation function. In this section, our investigation shifts to the state space division in general policy networks, which encompasses the bias $b^i$ and adopts $\ReLU$ as the activation function. Visualized state-action patterns for this network are shown in Fig. 11, where linear division lines are evident in the state space. However, for a general network, the linear division line is subject to unpredictability, as it remains uncertain whether the division line is perpendicular to weight vectors. While the aforementioned examples may not elucidate the factors affecting division line patterns, they unequivocally establish the prevalence of linear state space division in general policy networks.

\begin{figure}[!t]
	\centering
	\includegraphics[]{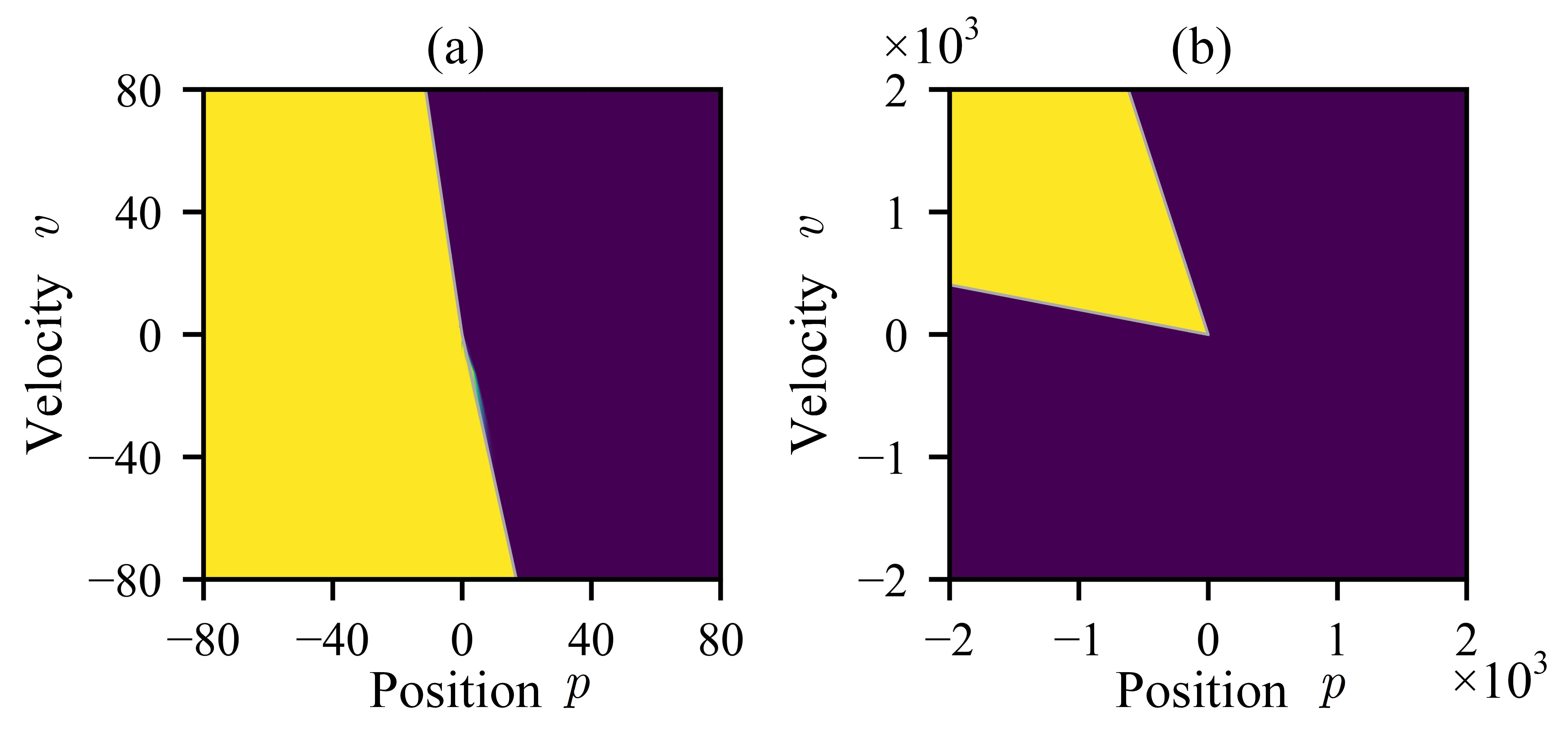}
	\caption{(a) Successful and (b) failed state-action patterns of general policy networks.}
\end{figure}

\section{Conclusion}
In this study, we uncover the emergence of linear state space division as policy networks expand across the entirety of space. Through a comprehensive analysis, this paper investigates the ramifications of this phenomenon on network performance, employing the double-integrator system as an illustrative example. By leveraging empirical data, this paper demonstrates that the linear division of the state space arises from the inherent saturability of the network and is independent of the optimization algorithms utilized. Furthermore, this paper illustrates how this linear division hampers a policy network's ability to accurately approximate the nonlinear ideal bang-bang control, resulting in inevitable overshooting. This revelation not only elucidates the reason for the network performance degradation in expanded state space but also implies the inherent limitations in the network's generalization capabilities. It can only achieve optimal control in specific scenarios while incurring significant performance degradation in others.

A current approach to addressing non-linearity degradation in distal state space involves generating a more conservative policy through the state regulation \cite{Han_clip}. This technique focuses on constraining the agent’s velocity within a predefined range. However, this approach compromises policy performance due to imposed velocity limits and the challenge still lies in the absence of an effective method to rectify the underlying network performance degradation due to the intrinsic structure of neural networks. 

In addition to devising techniques for enhancing network performance, our focus extends to evaluating the applicability of our analysis method within high-dimensional state space. Specifically, we are intrigued by the possibility of transitioning from a division line to a division hyperplane and the potential impact of this transition on the policy network performance. Moreover, how the degradation impact a nonlinear and high-dimension system will also be discussed in the future work.

\section{Acknowledgments}
This work was supported in part by the NNSFC\&CAAC under Grants
U2233209, in part by the Natural Science Foundation of Sichuan, China under Grant 2023NSFSC0484.

\bibliography{ref}

\begin{thebibliography}{60}
\providecommand{\natexlab}[1]{#1}

\bibitem[{Aractingi et~al.(2020)Aractingi, Dance, Perez, and
  Silander}]{regularization_1}
Aractingi, M.; Dance, C.; Perez, J.; and Silander, T. 2020.
\newblock Improving the Generalization of Visual Navigation Policies using
  Invariance Regularization.

\bibitem[{Bai et~al.(2021)Bai, Luo, Zhao, Wen, and Wang}]{adversarial_aug_3}
Bai, T.; Luo, J.; Zhao, J.; Wen, B.; and Wang, Q. 2021.
\newblock Recent Advances in Adversarial Training for Adversarial Robustness.
\newblock In \emph{Proceedings of the Thirtieth International Joint Conference
  on Artificial Intelligence, {IJCAI-21}}, 4312--4321. International Joint
  Conferences on Artificial Intelligence Organization.

\bibitem[{Bayerlein, De~Kerret, and Gesbert(2018)}]{dynamic_adjust_3}
Bayerlein, H.; De~Kerret, P.; and Gesbert, D. 2018.
\newblock Trajectory Optimization for Autonomous Flying Base Station via
  Reinforcement Learning.
\newblock In \emph{2018 IEEE 19th International Workshop on Signal Processing
  Advances in Wireless Communications (SPAWC)}, 1--5.

\bibitem[{Bengio et~al.(2009)Bengio, Louradour, Collobert, and
  Weston}]{curriculum_1}
Bengio, Y.; Louradour, J.; Collobert, R.; and Weston, J. 2009.
\newblock Curriculum Learning.
\newblock In \emph{Proceedings of the 26th Annual International Conference on
  Machine Learning}, 41–48. New York, NY, USA: Association for Computing
  Machinery.

\bibitem[{Chen and Li(2020)}]{robust_3}
Chen, S.; and Li, Y. 2020.
\newblock An Overview of Robust Reinforcement Learning.
\newblock In \emph{2020 IEEE International Conference on Networking, Sensing
  and Control (ICNSC)}, 1--6.

\bibitem[{Cobbe et~al.(2019)Cobbe, Klimov, Hesse, Kim, and
  Schulman}]{Ge_intro_2}
Cobbe, K.; Klimov, O.; Hesse, C.; Kim, T.; and Schulman, J. 2019.
\newblock Quantifying Generalization in Reinforcement Learning.
\newblock In Chaudhuri, K.; and Salakhutdinov, R., eds., \emph{Proceedings of
  the 36th International Conference on Machine Learning}, volume~97 of
  \emph{Proceedings of Machine Learning Research}, 1282--1289. PMLR.

\bibitem[{Cohan et~al.(2022)Cohan, Kim, Rolnick, and van~de Panne}]{divi_main}
Cohan, S.; Kim, N.~H.; Rolnick, D.; and van~de Panne, M. 2022.
\newblock Understanding the Evolution of Linear Regions in Deep Reinforcement
  Learning.
\newblock arXiv:2210.13611.

\bibitem[{Dhebar et~al.(2022)Dhebar, Deb, Nageshrao, Zhu, and
  Filev}]{Dhebar2022_Inter1_1}
Dhebar, Y.; Deb, K.; Nageshrao, S.; Zhu, L.; and Filev, D. 2022.
\newblock Toward Interpretable-AI Policies Using Evolutionary Nonlinear
  Decision Trees for Discrete-Action Systems.
\newblock \emph{IEEE Transactions on Cybernetics}, 1--13.

\bibitem[{Dobrevski and Skočaj(2020)}]{dynamic_adjust_1}
Dobrevski, M.; and Skočaj, D. 2020.
\newblock Adaptive Dynamic Window Approach for Local Navigation.
\newblock In \emph{2020 IEEE/RSJ International Conference on Intelligent Robots
  and Systems (IROS)}, 6930--6936.

\bibitem[{Du, Liu, and Hu(2019)}]{DT_1}
Du, M.; Liu, N.; and Hu, X. 2019.
\newblock Techniques for Interpretable Machine Learning.
\newblock \emph{Commun. ACM}, 63(1): 68–77.

\bibitem[{Ericsson et~al.(2022)Ericsson, Gouk, Loy, and
  Hospedales}]{self_representation_sz}
Ericsson, L.; Gouk, H.; Loy, C.~C.; and Hospedales, T.~M. 2022.
\newblock Self-Supervised Representation Learning: Introduction, advances, and
  challenges.
\newblock \emph{{IEEE} Signal Processing Magazine}, 39(3): 42--62.

\bibitem[{Fran{\c{c}}ois-Lavet et~al.(2018)Fran{\c{c}}ois-Lavet, Henderson,
  Islam, Bellemare, and Pineau}]{drl_intro3}
Fran{\c{c}}ois-Lavet, V.; Henderson, P.; Islam, R.; Bellemare, M.~G.; and
  Pineau, J. 2018.
\newblock An Introduction to Deep Reinforcement Learning.
\newblock \emph{Foundations and Trends{\textregistered} in Machine Learning},
  11(3-4): 219--354.

\bibitem[{Fujimoto, van Hoof, and Meger(2018)}]{TD3}
Fujimoto, S.; van Hoof, H.; and Meger, D. 2018.
\newblock Addressing Function Approximation Error in Actor-Critic Methods.
\newblock arXiv:1802.09477.

\bibitem[{Guo et~al.(2019)Guo, Shi, Kumar, Grauman, Rosing, and
  Feris}]{transfer2}
Guo, Y.; Shi, H.; Kumar, A.; Grauman, K.; Rosing, T.; and Feris, R. 2019.
\newblock SpotTune: Transfer Learning Through Adaptive Fine-Tuning.
\newblock In \emph{2019 IEEE/CVF Conference on Computer Vision and Pattern
  Recognition (CVPR)}, 4800--4809.

\bibitem[{Guo, Wu, and Lee(2022)}]{meta2}
Guo, Y.; Wu, Q.; and Lee, H. 2022.
\newblock Learning Action Translator for Meta Reinforcement Learning on
  Sparse-Reward Tasks.
\newblock In \emph{AAAI}, 6792--6800.

\bibitem[{Haarnoja et~al.(2019)Haarnoja, Zhou, Hartikainen, Tucker, Ha, Tan,
  Kumar, Zhu, Gupta, Abbeel, and Levine}]{SAC}
Haarnoja, T.; Zhou, A.; Hartikainen, K.; Tucker, G.; Ha, S.; Tan, J.; Kumar,
  V.; Zhu, H.; Gupta, A.; Abbeel, P.; and Levine, S. 2019.
\newblock Soft Actor-Critic Algorithms and Applications.
\newblock arXiv:1812.05905.

\bibitem[{Han et~al.(2023{\natexlab{a}})Han, Cheng, Xi, and
  Lv}]{simplified_han}
Han, H.; Cheng, J.; Xi, Z.; and Lv, M. 2023{\natexlab{a}}.
\newblock Symmetric actor–critic deep reinforcement learning for cascade
  quadrotor flight control.
\newblock \emph{Neurocomputing}, 559: 126789.

\bibitem[{Han et~al.(2023{\natexlab{b}})Han, Xi, Cheng, and Lv}]{Han_clip}
Han, H.; Xi, Z.; Cheng, J.; and Lv, M. 2023{\natexlab{b}}.
\newblock Obstacle Avoidance Based on Deep Reinforcement Learning and
  Artificial Potential Field.
\newblock In \emph{2023 9th International Conference on Control, Automation and
  Robotics (ICCAR)}, 215--220.

\bibitem[{Hanin and Rolnick(2019)}]{hanin2019_node}
Hanin, B.; and Rolnick, D. 2019.
\newblock Complexity of Linear Regions in Deep Networks.
\newblock arXiv:1901.09021.

\bibitem[{Hansen et~al.(2021)Hansen, Jangir, Sun, Alenyà, Abbeel, Efros,
  Pinto, and Wang}]{hansen2021selfsupervised}
Hansen, N.; Jangir, R.; Sun, Y.; Alenyà, G.; Abbeel, P.; Efros, A.~A.; Pinto,
  L.; and Wang, X. 2021.
\newblock Self-Supervised Policy Adaptation during Deployment.
\newblock arXiv:2007.04309.

\bibitem[{Hinton and Salakhutdinov(2006)}]{finetune_1}
Hinton, G.~E.; and Salakhutdinov, R.~R. 2006.
\newblock Reducing the Dimensionality of Data with Neural Networks.
\newblock \emph{Science}, 313(5786): 504--507.

\bibitem[{Hodge, Hawkins, and Alexander(2021)}]{example3264}
Hodge, V.~J.; Hawkins, R.; and Alexander, R. 2021.
\newblock Deep Reinforcement Learning for Drone Navigation Using Sensor Data.
\newblock \emph{Neural Comput. Appl.}, 33(6): 2015–2033.

\bibitem[{Hsu et~al.(2022)Hsu, Chen, Lu, Liu, and Yu}]{adversarial_aug_1}
Hsu, C.-Y.; Chen, P.-Y.; Lu, S.; Liu, S.; and Yu, C.-M. 2022.
\newblock Adversarial Examples Can Be Effective Data Augmentation for
  Unsupervised Machine Learning.
\newblock In \emph{AAAI}, 6926--6934.

\bibitem[{Hu et~al.(2022)Hu, Gao, Wan, Wang, and Zhai}]{curriculum_3}
Hu, Z.; Gao, X.; Wan, K.; Wang, Q.; and Zhai, Y. 2022.
\newblock Asynchronous Curriculum Experience Replay: A Deep Reinforcement
  Learning Approach for UAV Autonomous Motion Control in Unknown Dynamic
  Environments.
\newblock arXiv:2207.01251.

\bibitem[{Huang et~al.(2021)Huang, Guan, Xiao, and Lu}]{Domain_random_2}
Huang, J.; Guan, D.; Xiao, A.; and Lu, S. 2021.
\newblock FSDR: Frequency Space Domain Randomization for Domain Generalization.
\newblock In \emph{2021 IEEE/CVF Conference on Computer Vision and Pattern
  Recognition (CVPR)}, 6887--6898.

\bibitem[{Joshi, Virdi, and Chowdhary(2020)}]{simplified_2}
Joshi, G.; Virdi, J.; and Chowdhary, G. 2020.
\newblock Design and flight evaluation of deep model reference adaptive
  controller.
\newblock In \emph{AIAA Scitech 2020 Forum}, volume~6, 1--17.

\bibitem[{Julian et~al.(2020)Julian, Swanson, Sukhatme, Levine, Finn, and
  Hausman}]{finetune_3}
Julian, R.; Swanson, B.; Sukhatme, G.~S.; Levine, S.; Finn, C.; and Hausman, K.
  2020.
\newblock Never Stop Learning: The Effectiveness of Fine-Tuning in Robotic
  Reinforcement Learning.
\newblock arXiv:2004.10190.

\bibitem[{Kamath and Liu(2021)}]{post_hoc_2}
Kamath, U.; and Liu, J. 2021.
\newblock \emph{Post-Hoc Interpretability and Explanations}, 167--216.
\newblock Springer International Publishing.

\bibitem[{Karia and Srivastava(2022)}]{self_representation_1}
Karia, R.; and Srivastava, S. 2022.
\newblock Relational Abstractions for Generalized Reinforcement Learning on
  Symbolic Problems.
\newblock arXiv:2204.12665.

\bibitem[{Kirk et~al.(2023)Kirk, Zhang, Grefenstette, and
  Rocktäschel}]{Ge_intro_1}
Kirk, R.; Zhang, A.; Grefenstette, E.; and Rocktäschel, T. 2023.
\newblock A Survey of Zero-shot Generalisation in Deep Reinforcement Learning.
\newblock \emph{Journal of Artificial Intelligence Research}, 76: 201--264.

\bibitem[{Kirsch, van Steenkiste, and Schmidhuber(2020)}]{meta1Ge2}
Kirsch, L.; van Steenkiste, S.; and Schmidhuber, J. 2020.
\newblock Improving Generalization in Meta Reinforcement Learning using Learned
  Objectives.
\newblock arXiv:1910.04098.

\bibitem[{Kostrikov, Yarats, and Fergus(2021)}]{augmented_data_2}
Kostrikov, I.; Yarats, D.; and Fergus, R. 2021.
\newblock Image Augmentation Is All You Need: Regularizing Deep Reinforcement
  Learning from Pixels.
\newblock arXiv:2004.13649.

\bibitem[{Laskin et~al.(2020)Laskin, Lee, Stooke, Pinto, Abbeel, and
  Srinivas}]{augmented_data_1}
Laskin, M.; Lee, K.; Stooke, A.; Pinto, L.; Abbeel, P.; and Srinivas, A. 2020.
\newblock Reinforcement Learning with Augmented Data.
\newblock arXiv:2004.14990.

\bibitem[{Lee et~al.(2019)Lee, Lee, Shin, and Lee}]{Domain_random_4}
Lee, K.; Lee, K.; Shin, J.; and Lee, H. 2019.
\newblock Network Randomization: A Simple Technique for Generalization in Deep
  Reinforcement Learning.
\newblock In \emph{International Conference on Learning Representations}.

\bibitem[{Li(2018)}]{drl_intro}
Li, Y. 2018.
\newblock Deep Reinforcement Learning: An Overview.
\newblock arXiv:1701.07274.

\bibitem[{Lillicrap et~al.(2019)Lillicrap, Hunt, Pritzel, Heess, Erez, Tassa,
  Silver, and Wierstra}]{DDPG}
Lillicrap, T.~P.; Hunt, J.~J.; Pritzel, A.; Heess, N.; Erez, T.; Tassa, Y.;
  Silver, D.; and Wierstra, D. 2019.
\newblock Continuous control with deep reinforcement learning.
\newblock arXiv:1509.02971.

\bibitem[{Liu et~al.(2023)Liu, Liu, An, Gao, Yang, and Li}]{Liu2023_Inter1_3}
Liu, X.; Liu, S.; An, B.; Gao, Y.; Yang, S.; and Li, W. 2023.
\newblock Effective Interpretable Policy Distillation via Critical Experience
  Point Identification.
\newblock \emph{IEEE Intelligent Systems}, 1--10.

\bibitem[{Morimoto and Doya(2005)}]{robust_sz}
Morimoto, J.; and Doya, K. 2005.
\newblock {Robust Reinforcement Learning}.
\newblock \emph{Neural Computation}, 17(2): 335--359.

\bibitem[{Packer et~al.(2019)Packer, Gao, Kos, Krähenbühl, Koltun, and
  Song}]{packer2019assessing}
Packer, C.; Gao, K.; Kos, J.; Krähenbühl, P.; Koltun, V.; and Song, D. 2019.
\newblock Assessing Generalization in Deep Reinforcement Learning.
\newblock arXiv:1810.12282.

\bibitem[{Park and Park(2022)}]{curriculum_2}
Park, J.; and Park, K. 2022.
\newblock Indoor Path Planning for Multiple Unmanned Aerial Vehicles via
  Curriculum Learning.
\newblock In \emph{2022 13th International Conference on Information and
  Communication Technology Convergence (ICTC)}, 1238--1241. {IEEE}.

\bibitem[{Pascanu, Montufar, and Bengio(2014)}]{pascanu2014_count2}
Pascanu, R.; Montufar, G.; and Bengio, Y. 2014.
\newblock On the number of response regions of deep feed forward networks with
  piece-wise linear activations.
\newblock arXiv:1312.6098.

\bibitem[{Peng et~al.(2018)Peng, Andrychowicz, Zaremba, and
  Abbeel}]{Domain_random_3}
Peng, X.~B.; Andrychowicz, M.; Zaremba, W.; and Abbeel, P. 2018.
\newblock Sim-to-Real Transfer of Robotic Control with Dynamics Randomization.
\newblock In \emph{2018 {IEEE} International Conference on Robotics and
  Automation ({ICRA})}.

\bibitem[{Raileanu et~al.(2021)Raileanu, Goldstein, Yarats, Kostrikov, and
  Fergus}]{augmented_data_3}
Raileanu, R.; Goldstein, M.; Yarats, D.; Kostrikov, I.; and Fergus, R. 2021.
\newblock Automatic Data Augmentation for Generalization in Deep Reinforcement
  Learning.
\newblock arXiv:2006.12862.

\bibitem[{Rajeswaran et~al.(2017)Rajeswaran, Ghotra, Ravindran, and
  Levine}]{robust_2}
Rajeswaran, A.; Ghotra, S.; Ravindran, B.; and Levine, S. 2017.
\newblock EPOpt: Learning Robust Neural Network Policies Using Model Ensembles.
\newblock arXiv:1610.01283.

\bibitem[{Rusu et~al.(2022)Rusu, Rabinowitz, Desjardins, Soyer, Kirkpatrick,
  Kavukcuoglu, Pascanu, and Hadsell}]{finetune_2}
Rusu, A.~A.; Rabinowitz, N.~C.; Desjardins, G.; Soyer, H.; Kirkpatrick, J.;
  Kavukcuoglu, K.; Pascanu, R.; and Hadsell, R. 2022.
\newblock Progressive Neural Networks.
\newblock arXiv:1606.04671.

\bibitem[{Schulman et~al.(2017)Schulman, Wolski, Dhariwal, Radford, and
  Klimov}]{PPO}
Schulman, J.; Wolski, F.; Dhariwal, P.; Radford, A.; and Klimov, O. 2017.
\newblock Proximal Policy Optimization Algorithms.
\newblock arXiv:1707.06347.

\bibitem[{Serra, Tjandraatmadja, and Ramalingam(2018)}]{serra2018_count1}
Serra, T.; Tjandraatmadja, C.; and Ramalingam, S. 2018.
\newblock Bounding and Counting Linear Regions of Deep Neural Networks.
\newblock arXiv:1711.02114.

\bibitem[{Slaoui et~al.(2020)Slaoui, Clements, Foerster, and
  Toth}]{regularization_2}
Slaoui, R.~B.; Clements, W.~R.; Foerster, J.~N.; and Toth, S. 2020.
\newblock Robust Visual Domain Randomization for Reinforcement Learning.
\newblock arXiv:1910.10537.

\bibitem[{Soares et~al.(2021)Soares, Angelov, Costa, Castro, Nageshrao, and
  Filev}]{Soares2020_Inter1_2}
Soares, E.; Angelov, P.~P.; Costa, B.; Castro, M. P.~G.; Nageshrao, S.; and
  Filev, D. 2021.
\newblock Explaining Deep Learning Models Through Rule-Based Approximation and
  Visualization.
\newblock \emph{IEEE Transactions on Fuzzy Systems}, 29(8): 2399--2407.

\bibitem[{Sun et~al.(2022)Sun, Greene, Le, Bell, Chowdhary, and
  Dixon}]{simplified_1}
Sun, R.; Greene, M.~L.; Le, D.~M.; Bell, Z.~I.; Chowdhary, G.; and Dixon, W.~E.
  2022.
\newblock Lyapunov-Based Real-Time and Iterative Adjustment of Deep Neural
  Networks.
\newblock \emph{IEEE Control Systems Letters}, 193--198.

\bibitem[{Tobin et~al.(2017)Tobin, Fong, Ray, Schneider, Zaremba, and
  Abbeel}]{data_randomization_tobin}
Tobin, J.; Fong, R.; Ray, A.; Schneider, J.; Zaremba, W.; and Abbeel, P. 2017.
\newblock Domain Randomization for Transferring Deep Neural Networks from
  Simulation to the Real World.
\newblock arXiv:1703.06907.

\bibitem[{Volpi et~al.(2018)Volpi, Namkoong, Sener, Duchi, Murino, and
  Savarese}]{adversarial_aug_2}
Volpi, R.; Namkoong, H.; Sener, O.; Duchi, J.~C.; Murino, V.; and Savarese, S.
  2018.
\newblock Generalizing to Unseen Domains via Adversarial Data Augmentation.
\newblock In Bengio, S.; Wallach, H.; Larochelle, H.; Grauman, K.;
  Cesa-Bianchi, N.; and Garnett, R., eds., \emph{Advances in Neural Information
  Processing Systems}, volume~31. Curran Associates, Inc.

\bibitem[{Vrbančič and Podgorelec(2020)}]{transfer1}
Vrbančič, G.; and Podgorelec, V. 2020.
\newblock Transfer Learning With Adaptive Fine-Tuning.
\newblock \emph{IEEE Access}, 8: 196197--196211.

\bibitem[{Wang et~al.(2018)Wang, Wang, Xu, and Zhang}]{bang_bang_control}
Wang, G.; Wang, L.; Xu, Y.; and Zhang, Y. 2018.
\newblock \emph{Time Optimal Control of Evolution Equations}, chapter~6.
\newblock Cham, Switzerland: Birkhäuser.

\bibitem[{WANG et~al.(2020)WANG, Kang, Shao, and Feng}]{mixreg}
WANG, K.; Kang, B.; Shao, J.; and Feng, J. 2020.
\newblock Improving Generalization in Reinforcement Learning with Mixture
  Regularization.
\newblock In Larochelle, H.; Ranzato, M.; Hadsell, R.; Balcan, M.; and Lin, H.,
  eds., \emph{Advances in Neural Information Processing Systems}, volume~33,
  7968--7978. Curran Associates, Inc.

\bibitem[{Wang et~al.(2022)Wang, Wang, Liang, Zhao, Huang, Xu, Dai, and
  Miao}]{drl_intro2}
Wang, X.; Wang, S.; Liang, X.; Zhao, D.; Huang, J.; Xu, X.; Dai, B.; and Miao,
  Q. 2022.
\newblock Deep Reinforcement Learning: A Survey.
\newblock \emph{IEEE Transactions on Neural Networks and Learning Systems},
  1--15.

\bibitem[{Wu et~al.(2022)Wu, Wu, Chen, Xu, and Li}]{Ge6}
Wu, K.; Wu, M.; Chen, Z.; Xu, Y.; and Li, X. 2022.
\newblock Generalizing Reinforcement Learning through Fusing Self-Supervised
  Learning into Intrinsic Motivation.
\newblock In \emph{AAAI}, 8683--8690.

\bibitem[{Xin et~al.(2017)Xin, Zhao, Liu, and Li}]{dynamic_adjust_2}
Xin, J.; Zhao, H.; Liu, D.; and Li, M. 2017.
\newblock Application of deep reinforcement learning in mobile robot path
  planning.
\newblock In \emph{2017 Chinese Automation Congress (CAC)}, 7112--7116.

\bibitem[{Yang et~al.(2022)Yang, Sim\~{a}o, Tindemans, and Spaan}]{safe2}
Yang, Q.; Sim\~{a}o, T.~D.; Tindemans, S.~H.; and Spaan, M. T.~J. 2022.
\newblock Safety-Constrained Reinforcement Learning with a Distributional
  Safety Critic.
\newblock \emph{Mach. Learn.}, 112(3): 859–887.

\bibitem[{Yang et~al.(2021)Yang, Simão, Tindemans, and Spaan}]{safe1}
Yang, Q.; Simão, T.~D.; Tindemans, S.~H.; and Spaan, M. T.~J. 2021.
\newblock WCSAC: Worst-Case Soft Actor Critic for Safety-Constrained
  Reinforcement Learning.
\newblock \emph{Proceedings of the AAAI Conference on Artificial Intelligence},
  35(12): 10639--10646.

\end{thebibliography}

\end{document}